\documentclass[journal]{IEEEtran}

\usepackage{cite}
\usepackage{epsfig}

\usepackage{amsmath}
\usepackage{amssymb}
\usepackage{color}
\usepackage{url}
\usepackage{makecell}
\usepackage{verbatim}
\usepackage{multirow}
\usepackage{tabularx}
\usepackage{subfigure}

\usepackage[colorlinks,linkcolor=red]{hyperref}

\usepackage{autobreak}
\usepackage{breqn}

\usepackage{booktabs} %

\usepackage{soul}
\soulregister{\cite}7 %
\soulregister{\citep}7 %
\soulregister{\citet}7 %
\soulregister{\ref}7 %
\soulregister{\pageref}7 %

\usepackage[table,x11names]{xcolor}

\ifCLASSINFOpdf
\else
\fi

\begin{document}

\title{Semantic Face Compression for Metaverse: A Compact 3D Descriptor Based Approach}

\author{Binzhe Li, 
        Bolin Chen, 
        Zhao Wang,
        Shiqi Wang,~\IEEEmembership{Senior Member,~IEEE}, \\
        and Yan Ye,~\IEEEmembership{Senior Member,~IEEE} 
\thanks{Binzhe Li,  Bolin Chen and Shiqi Wang are with the Department of Computer Science, City University of Hong Kong,
Hong Kong (e-mail: binzheli2-c@my.cityu.edu.hk; bolinchen3-c@my.cityu.edu.hk; shiqwang@cityu.edu.hk).}
\thanks{Zhao Wang is with the Peking University, Beijing, 100871, China. (e-mail: zhaowang@pku.edu.cn).}
\thanks{Yan Ye is with Alibaba Group U.S., Sunnyvale, CA 94085 USA (e-mail: yan.ye@alibaba-inc.com).} 
}

\markboth{submitted to IEEE Transactions on Circuits and Systems for Video Technology}{Right}
 
\maketitle
 
\begin{abstract}
In this letter, we envision a new metaverse communication paradigm for virtual avatar faces, and develop the semantic face compression with compact 3D facial descriptors. The fundamental principle is that the communication of virtual avatar faces primarily emphasizes the conveyance of semantic information. In light of this, the proposed scheme offers the advantages of being highly flexible, efficient and semantically meaningful. The semantic face compression, which allows the communication of the descriptors for artificial intelligence based understanding, could facilitate numerous applications without the involvement of humans in metaverse. The promise of the proposed paradigm is also demonstrated by performance comparisons with the state-of-the-art video coding standard, Versatile Video Coding. A significant  improvement in terms of rate-accuracy performance has been achieved. The proposed scheme is expected to enable numerous applications, such as digital human communication based on machine analysis, and to form the cornerstone of interaction and communication in the metaverse.
\end{abstract}


\IEEEpeerreviewmaketitle

\section{Introduction}

\IEEEPARstart{R}{ecently}, there has been a tremendous demand for high-efficiency interactive face communications, coinciding with the popularization of the metaverse. In the metaverse, face communication between virtual avatars serves the purpose of conveying semantic information, as the visual information cannot be physically perceived by the virtual avatars' ``eyes''. In this context, the term ``face" represents the virtual mapping of the physical world, specifically referring to the representation of the physical face for virtual avatars. As such, the artificial intelligence (AI) systems which empower the virtual avatars are treated as the ultimate receiver. This particular application scenario presents unique challenges for face communication systems. Existing solutions based on face video compression have certain limitations, as they are primarily developed with the purpose of signal-level reconstruction, without fully considering the interactive and intelligent tasks required in the metaverse.

The advancement of video coding is driven by the development of video coding standards, such as H.264/AVC~\cite{wiegand2003overview}, H.265/HEVC~\cite{sullivan2012overview}, and H.266/VVC~\cite{VVC} and AVS~\cite{ma2022evolutionAVS}. Recently, end-to-end learning-based coding algorithms have been proposed to achieve competitive performance, such as Deep Video Compression framework (DVC)~\cite{lu2019dvc}, Multiple frames prediction for Learned Video Compression (M-LVC)~\cite{MLVC} and Decomposed Motion modeling for learned Video Compression (DMVC)~\cite{lin2022dmvc}. Though promising coding performance has been achieved, there are still challenges ahead in low bit-rate face video communications. To address these challenges, the temporal variations of face have been characterized by first-order motion model (FOMM)~\cite{FOMM}, Face\_vid2vid~\cite{wang2021Nvidia}, and compact feature vector~\cite{chen2023compact}, all of which possess many favorable properties. Though promising rate-distortion (RD) performance has been achieved with these methods, the temporal movement information extracted and compactly represented by the learned networks may not accurately and straightforwardly represent face semantics. In light of this, we resort to the utilization of 3D descriptors, which are compact, semantically meaningful, and flexible, based on the prior 3D modeling of the face. 

\begin{figure}[tb]
\centering
\includegraphics[width=0.5\textwidth]{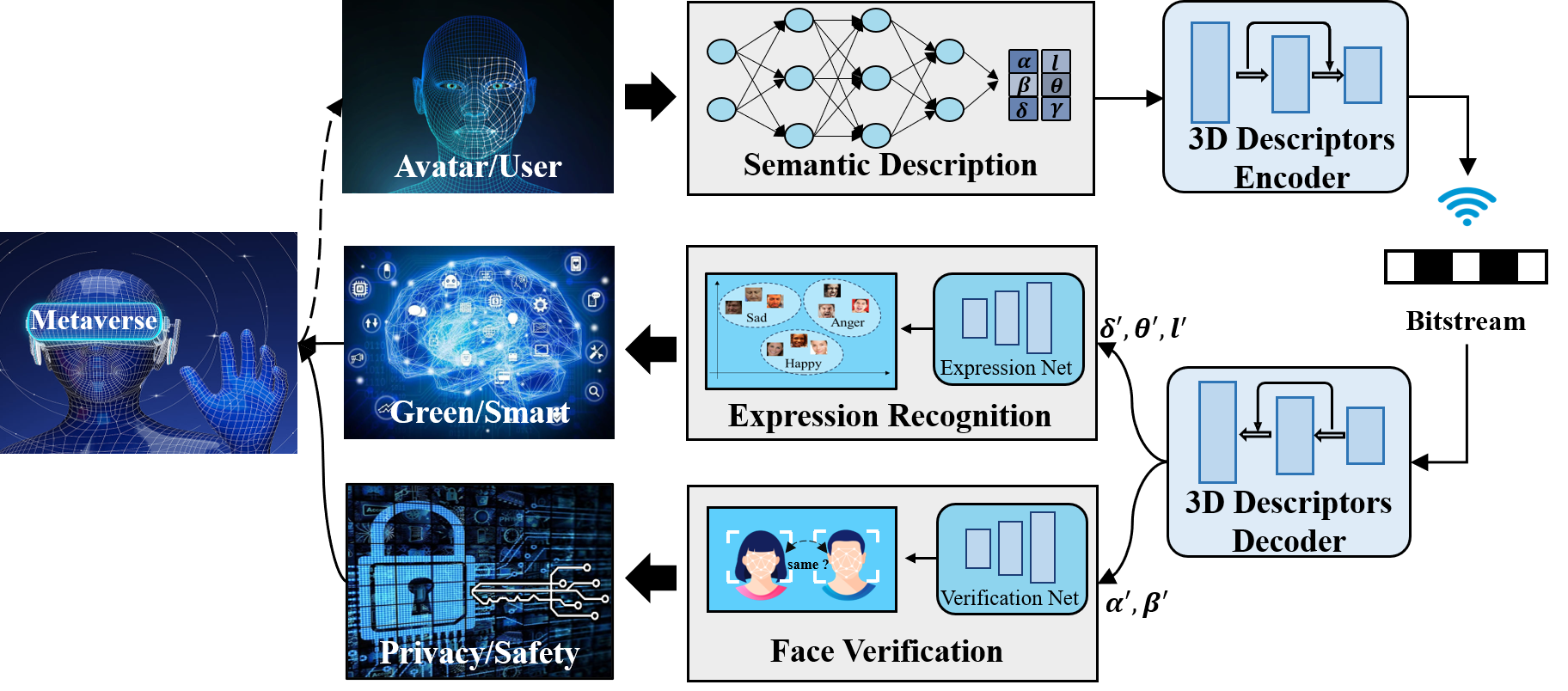}
\caption{ The paradigm of the proposed descriptors-based communication framework.}
\label{fig_paradigm}
\end{figure}

In this letter, we propose a viable solution of semantic face compression for metaverse, as shown in Fig.~\ref{fig_paradigm}. In particular, we first propose a  face compression framework that leverages the 3D face prior and descriptors for semantic communication in metaverse. Subsequently, we show how the descriptors that represent the 3D face can be feasibly transmitted based on the end-to-end compression network. Finally, we introduce the utility of such framework towards intelligent applications, targeting at better understanding the face without reconstructing the signal. The proposed scheme enjoys the advantages of high flexibility and compact representation due to the semantic level representation, leading to better rate-accuracy (RA) performance. Furthermore, based on face descriptors and their semantics, the framework allows friendly and more direct interactions with users, which leads to great potential for metaverse communication.

\begin{figure*}[t]
\centering
\includegraphics[width=0.95\textwidth]{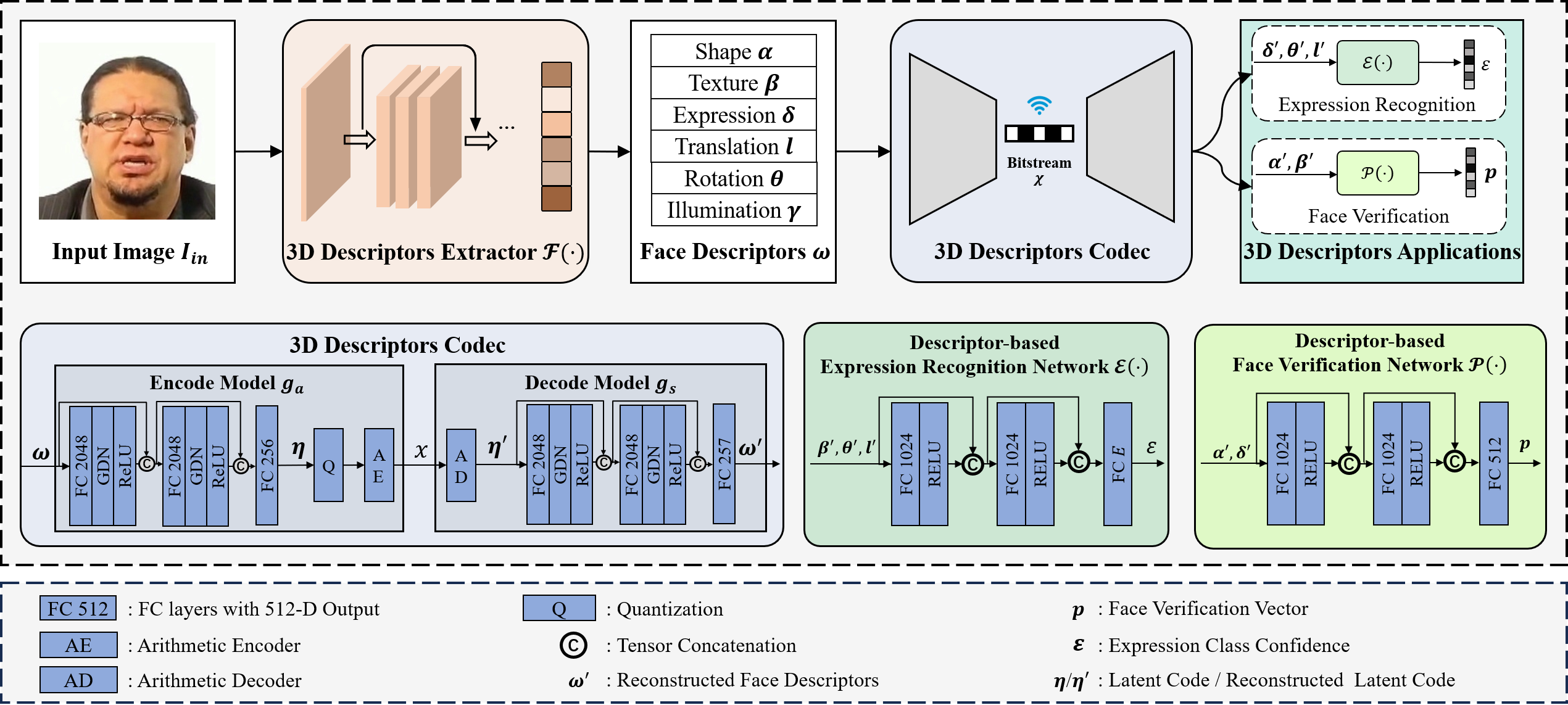}
\caption{ Illustration of the proposed face semantic compression framework, which consists of 3D descriptors extraction, compression, and face expression recognition and verification based on the 3D descriptors. }
\label{fig_framework}
\end{figure*}

\section{The Proposed Semantic Face Compression Scheme}

\subsection{Framework in a Nutshell}

The proposed framework is dedicated to the communication of faces among virtual avatars which are empowered by AI. As illustrated in Fig.~\ref{fig_paradigm}, the design philosophy is the conveyance of 3D descriptors. There are three reasons behind the proposed framework. First, the 3D descriptors are not only compact but also encompass explicit semantic meanings. Second, the mature 3D Morphable Models (3DMM)~\cite{blanz1999morphable} can serve as the facial priors for such semantic compression, and such context could further make the semantic compression more efficient. Third, the face descriptors could be directly utilized for semantic understanding tasks based on AI. 

As shown in Fig.~\ref{fig_framework}, the proposed framework is mainly composed of the 3D descriptor extractor, the 3D descriptor codec, and the intelligent applications based on received 3D descriptors. It serves as a holistic solution for face semantic communication in metaverse based on compact 3D face descriptor representation~\cite{li2022towards}. Specifically,  given the input image $I_{in}$, the descriptor extractor $\mathcal{F}\left ( \cdot  \right ) $ is employed to obtain the descriptors $\boldsymbol{\omega}$, which are mainly composed of expression $\boldsymbol{\delta }$, rotation $\boldsymbol{\theta}$, translation $\boldsymbol{l}$, illumination $\boldsymbol{\gamma}$, shape $\boldsymbol{\alpha}$ and texture $\boldsymbol{\beta}$. After that, these semantic descriptors $\boldsymbol{\omega}$ are compressed by the 3D descriptor codec, where $\boldsymbol{\omega}$ are compactly represented as the bitstream $\boldsymbol{\chi}$ with the encoder and reconstructed by the decoder as the reconstructed descriptors $\boldsymbol{\omega}^\prime$. Finally, the reconstructed descriptors facilitate two machine vision tasks: the descriptor-based expression recognition task and the descriptor-based face verification task. For the descriptor-based expression recognition task, given the reconstructed $\boldsymbol{\delta}^\prime$, $\boldsymbol{\theta}^\prime$ and  $\boldsymbol{l}^\prime$ descriptors, a descriptor-based expression recognition network $\mathcal{E}\left ( \cdot  \right ) $ is developed to infer the expression confidence $\boldsymbol{\varepsilon}$ and achieve semantic-level expression classifications. Regarding the descriptor-based face verification task, the proposed verification network $\mathcal{P}\left ( \cdot  \right ) $ can transform the reconstructed $\boldsymbol{\alpha}^\prime$ and  $\boldsymbol{\beta}^\prime$ descriptors into a specific face verification vector $\boldsymbol{p} \in \mathbb{R}^{512} $, thus facilitating the face verification and privacy protection.

Overall, the proposed face semantic compression framework could support the face communication between virtual avatars in the metaverse. On one hand, the utilization of received face information for virtual avatars is semantic understanding instead of signal-level perception. Our proposed framework aligns this underlying principle as the compact descriptors can directly support the intelligent tasks. On the other hand, the interactivity can be further enhanced with the proposed framework, as the conveyed descriptors can even be manipulated to align with the requirement from the end-user. 

\subsection{3D Descriptor Extraction}

In this subsection, the semantic descriptors are extracted from the input image. In particular, we use the 3DMM~\cite{blanz1999morphable} parameters as face descriptors. The descriptors which are denoted as $\boldsymbol{\omega}$ is obtained by,
\begin{equation}
    \boldsymbol{\omega} = \mathcal{F}\left (  {I}_{in}  \right ),
\end{equation}
where $\mathcal{F}\left ( \cdot  \right ) $ means the extractor network, and ${I}_{in}$ is input image.
Specifically,  $\boldsymbol{\omega}$ combines a collection of representation parameters,
\begin{equation}
    \boldsymbol{\omega} = \left \{  \boldsymbol{\alpha}, \boldsymbol{\beta},   \boldsymbol{\delta}, \boldsymbol{\theta},  \boldsymbol{l},  \boldsymbol{\gamma} \right \}.
\end{equation}
In detail, $\boldsymbol{\omega} \in \mathbb{R}^{257}$ mainly includes shape descriptors $\boldsymbol{\alpha} \in \mathbb{R}^{80}$, texture descriptors $\boldsymbol{\beta} \in \mathbb{R}^{80} $, expression descriptors $\boldsymbol{\delta}  \in \mathbb{R}^{64} $, rotation angle descriptors $\boldsymbol{\theta}  \in \mathbb{R}^{3}$, translation descriptors $\boldsymbol{l} \in \mathbb{R}^{3}$, and illumination descriptors $\boldsymbol{\gamma} \in \mathbb{R}^{27}$. These descriptors contain semantic information for machine tasks. The pre-trained face reconstruction model Weakly-Supervised Multi-Face 3D Reconstruction (WM3DR)~\cite{Zhang2021WeaklySupervisedM3} is employed as the extractor to obtain these descriptors. 

\subsection{3D Descriptor Compression}
In this subsection, 3D face descriptors are compactly represented to facilitate high efficiency communication. We introduce the end-to-end descriptors compression, which converts the descriptors into latent code for compact representation, and the descriptors are reconstructed with the latent code at the decoder side. Specifically, to facilitate the transmission, the end-to-end compression framework consisting of the encoder $g_a$ and decoder $g_s$ can represent the 3D face descriptors $\boldsymbol{\omega}$ into the compact bitstream $\boldsymbol{\chi}$ and achieve the reconstruction of face descriptors $\boldsymbol{\omega}^\prime$. Herein, the encoding and decoding processes can be formulated as,
\begin{equation}
\begin{aligned}
\boldsymbol{\chi} =g_a\left(\boldsymbol{\omega}\right), ~
\boldsymbol{\omega}^\prime =g_s\left(\boldsymbol{\chi}\right).
\end{aligned}
\label{eq5}
\end{equation}

The encoder and decoder powered by the end-to-end deep neural networks are illustrated in Fig.~\ref{fig_framework}. Herein, $Q$ denotes the quantization operation, and $AE$ and $AD$ are the arithmetic encoder and decoder, respectively. In particular, the given extracted descriptors $\boldsymbol{\omega}$ for the current frame $I_{in}$ could be encoded to form the latent code $\boldsymbol{\eta}$ by the end-to-end learned network. As rounding quantization is usually not differential, we replace rounding operation for \(\boldsymbol{\eta} \) with adding uniform noise during training~\cite{balle2017end,balle2018variational}. For $AE$ and $AD$, the principle of zero-order exponential Golomb algorithm~\cite{ExponentialGolomb} is facilitated to transform $\boldsymbol{\omega}$ into binary codes, and the Prediction by Partial Matching algorithm~\cite{PPM} is also deployed to generate the final bitstream $\boldsymbol{\chi}$. 

More specifically, the network is designed to transform the input vector to the corresponding compact representations following the auto-encoder style network, and the ResNet-style shortcuts are also applied for better encoding performance. The network consists of a series of fully connected (FC) layers and Rectified Linear Unit (ReLU) layers~\cite{2011Deep}. Besides, the Generalized Divisive Normalization (GDN)~\cite{balle2017end,balle2018variational} is adopted as the nonlinear transform. The output dimension of the latent code is set to 256. 

During training, the loss function is composed of the rate loss $\mathcal{L}_R$ and mean absolute error (MAE) loss $\mathcal{L}_{MAE}$. In particular, the rate loss $\mathcal{L}_R$ is estimated with the entropy model~\cite{balle2017end, balle2018variational}. The MAE loss identifies the distortion between the $ \boldsymbol{\omega}$  and $ \boldsymbol{\omega}^\prime$.
As such, the final loss function is given by, 
\begin{equation}
\mathcal{L}_{codec}=\lambda_{mae}\mathcal{L}_{MAE}(\boldsymbol{\omega},\boldsymbol{\omega}^\prime)+\lambda_{R}\mathcal{L}_R(\boldsymbol{\chi}),
\label{eq6}
\end{equation}
where $\lambda_{mae}$ and $\lambda_{R}$ are the hyper-parameters, which are set as $\lambda_{mae}=1.0$ and $\lambda_{R}=0.001$. 

\subsection{Semantic Face Understanding Based on the Proposed Framework}
\label{ER}

In this subsection, instead of face understanding with the image as the input, two descriptors based semantic face understanding networks are developed for the proposed framework. In detail, a facial expression understanding network and a learning-based face verification network are proposed for expression recognition and face verification tasks, respectively. 

\subsubsection{Facial Expression Understanding}
For the facial expression understanding, a facial expression recognition network is proposed to estimate the facial expression with the descriptor as input. It is generally acknowledged that the shape, texture, and illumination are not related to the person’s expression. Therefore, only the reconstructed expression $\boldsymbol{\delta}^\prime$, angle $\boldsymbol{\theta}^\prime$, and translation $\boldsymbol{l}^\prime$ are taken for inference, and then the expression class confidence is obtained as,
\begin{equation}
\label{ }
 \boldsymbol{\varepsilon} =  \mathcal{E} \left ( \boldsymbol{\delta}^\prime, \boldsymbol{\theta}^\prime,  \boldsymbol{l}^\prime \right ) ,
\end{equation}
where $\mathcal{E}\left ( \cdot  \right ) $ denotes the facial expression recognition network and the expression class confidence $\boldsymbol{\varepsilon} \in \mathbb{R}^{E}$, with $E$ being the number of expression classes. Specifically, the network consists of a series of FC layers and ReLU layers. The ResNet-style shortcuts~\cite{ResNet} are applied between the FC layers, as shown in Fig.~\ref{fig_framework}. In training, the loss function is the cross-entropy loss for the classification task. Therefore, this network can perform facial expression recognition tasks based solely on semantic descriptors.

\subsubsection{Face Verification}
For the face verification task, a descriptor-based face verification network is developed to obtain the face verification vector that performs the verification task. Analogous to the FaceNet method~\cite{FaceNet}, the descriptor-based face verification method encourages one identity to be projected onto a single point in the verification vector space. Specifically, because the expression, translation, rotation, and illumination are unrelated to the identity, only the shape $\boldsymbol{\alpha}^\prime$ and texture $\boldsymbol{\beta}^\prime$ are taken as input. The output is the verification vector $\boldsymbol{p}$. The verification process is represented by,
\begin{equation}
\label{CEloss}
\boldsymbol{p} =  \mathcal{P}\left ( \boldsymbol{\alpha}^\prime, \boldsymbol{\beta}^\prime  \right )  \in \mathbb{R}^{K}, 
\end{equation}
where  $\mathcal{P}\left ( \cdot  \right ) $ corresponds to the  face verification network and the ${K}$ is the dimension of the output verification vector. Furthermore, the network also consists of FC layers and ReLU layers. The ResNet-style shortcuts~\cite{ResNet} are also applied, as shown in Fig.~\ref{fig_framework}. In training, the loss function is the mean square error (MSE) loss. We take the  FaceNet~\cite{FaceNet} 512-D verification vector as the ground truth. After obtaining the verification vectors, the face verification task becomes finding the distance threshold between the two verification vectors.

\section{Experimental Results}

\subsection{Experimental Setup}

\subsubsection{Descriptor Compression}
To train the descriptors compression network, the VoxCeleb dataset~\cite{VoxCeleb17} is introduced as the training set. In detail, one frame is randomly selected during training and is converted to the descriptor based on the 3D descriptor extractor~\cite{Zhang2021WeaklySupervisedM3}.  Then, these descriptors are used as training data. These data include different variations for pose and expression, such that they are adopted for network training. Regarding implementation details, we train the network using the Adam~\cite{Adam} optimizer with the 0.0001 learning rate, 64  batch size, and the parameters ($\beta_1 = 0.5$ and $\beta_2=0.999$). The epoch is set to 40, and the model training is conducted on the NVIDIA TESLA V100 GPU.

\subsubsection{Face Expression Recognition}
\label{ERExperimentalSetup}

For the training of descriptor-based expression recognition network, the facial emotion dataset in the wild with eight classes of expression labels (AffectNet-8)~\cite{mollahosseini2017affectnet} is introduced as the training and testing set. There are 210K images in the training dataset and  4K images in the testing dataset. In detail, these images in the AffectNet-8 are cropped and resized to 224$\times$224 pixels. The 3D descriptor extractor~\cite{Zhang2021WeaklySupervisedM3} converts the images to the descriptors which are taken as the inputs of the descriptor-based expression inference network. In training, the settings are the same as those of descriptor compression network training except that the epoch is set to 30.

To better verify the performance of our proposed scheme, we take the reference software of the state-of-the-art coding standard VVC (VTM-12.0)~\cite{VVC} as the baseline. More specifically, the RGB images are converted to YUV420 format files, which are further compressed by the VVC codec with the Quantization Parameters (QPs) 49, 51, and 53 to match the rates of proposed method. Finally, a multi-head cross-attention facial expression recognition network (i.e., DAN~\cite{wen2023distract}) is deployed to perform the expression recognition task with these reconstructed images. As for our proposed descriptor-based framework, face images can be represented as semantic descriptors via the 3D descriptor extractor and compressed into compact bitstream via the 3D descriptor codec. Afterwards, the reconstructed 3D descriptors are further fed into the descriptor-based expression recognition network. Herein, the recognition accuracy and coding bit consumption are both regarded as the performance evaluation measures.

\subsubsection{Face Verification}

\begin{table}[t]
\caption{ Performance comparisons for the expression recognition task.} 
\begin{center}
\begin{tabular}{m{50pt}<{\centering} m{37pt}<{\centering} m{33pt}<{\centering} m{33pt}<{\centering}  m{33pt}<{\centering} }
\hline
\specialrule{0em}{1pt}{1pt}
Performance Measures          &  {Proposed Framework}  &  {Baseline (QP=49)}  &  {Baseline (QP=51)} &  {Baseline (QP=53)}  \\ \hline

 Coding Bits & 1450.4               & 2117                  & 1782.7                               & 1519.4                              \\
\specialrule{0em}{1pt}{1pt}
Val Accuracy      & 47.49\%                & 49.03\%                & 44.91\%                              & 38.60\%                              \\ \hline
\end{tabular}
\end{center}
\label{ERPerformance}
\end{table}

\begin{table}[t]
\caption{ Performance comparisons for the face verification task.} 
\begin{center}

\begin{tabular}{m{50pt}<{\centering} m{37pt}<{\centering} m{33pt}<{\centering} m{33pt}<{\centering}  m{33pt}<{\centering} }
\hline
Performance Measures  &  {Proposed Framework}  &  {Baseline (QP=47)} &  {Baseline (QP=49)} &  {Baseline (QP=51)} \\ \hline
\specialrule{0em}{1pt}{1pt}
Coding Bits & 1409.67                     &     1940.47         & 1653.1                          & 1426.12                            \\
\specialrule{0em}{1pt}{1pt}
Val Accuracy      & 90.35\%                      &  89.38\%                & 85.16\%                              & 79.70\%                              \\ \hline
\end{tabular}
\end{center}
\label{FEPerformance}
\end{table}

Herein, CASIA-WebFace~\cite{yi2014learning} and Labeled Faces in the Wild (LFW) datasets~\cite{huang2008labeled} are used for training and testing, respectively. To align the images, the Multi-task Cascaded Convolutional Networks (MTCNN)~\cite{zhang2016joint} is introduced to detect the face region of input images, where face-centric images are cropped at the resolution of 160$\times$160 in the training and testing datasets. Afterwards, 3D descriptors extracted from processed face datasets are further converted to 512-D verification vectors via the FaceNet~\cite{FaceNet}. The training settings of the face verification network are also the same as the training of the face expression recognition network. 

In the evaluation, the baseline method and proposed method are compared. For the baseline method, the VVC codec is used to compress the testing images with QP 47, 49, and 51. Then, the FaceNet~\cite{FaceNet} extracts the verification vectors of these reconstructed images from VVC. In the proposed descriptor-based framework, the reconstructed descriptors are used to infer the verification vectors. As such, the accuracy of the face verification vectors can be finally obtained.

\subsection{Performance Evaluations}

This subsection shows the performance comparisons of two different frameworks for two machine tasks. Table~\ref{ERPerformance} illustrates that the proposed framework achieves advantageous RA performance over the baseline in the expression recognition task. In particular, the proposed descriptor-based framework shows a $8.89\%$ expression recognition accuracy gain compared to the baseline framework with even lower coding bits. As shown in Table~\ref{FEPerformance}, the proposed framework obtains a better RA performance than the baseline framework in the face verification task, with the $10.65\%$ verification accuracy improvement compared to the baseline framework at similar coding bits. Therefore, the proposed framework is superior to the baseline framework with the state-of-the-art codec.

\begin{figure}[t]
\centering
\begin{minipage}[b]{.47\linewidth}
  \centering
  \centerline{\includegraphics[width=4.3cm]{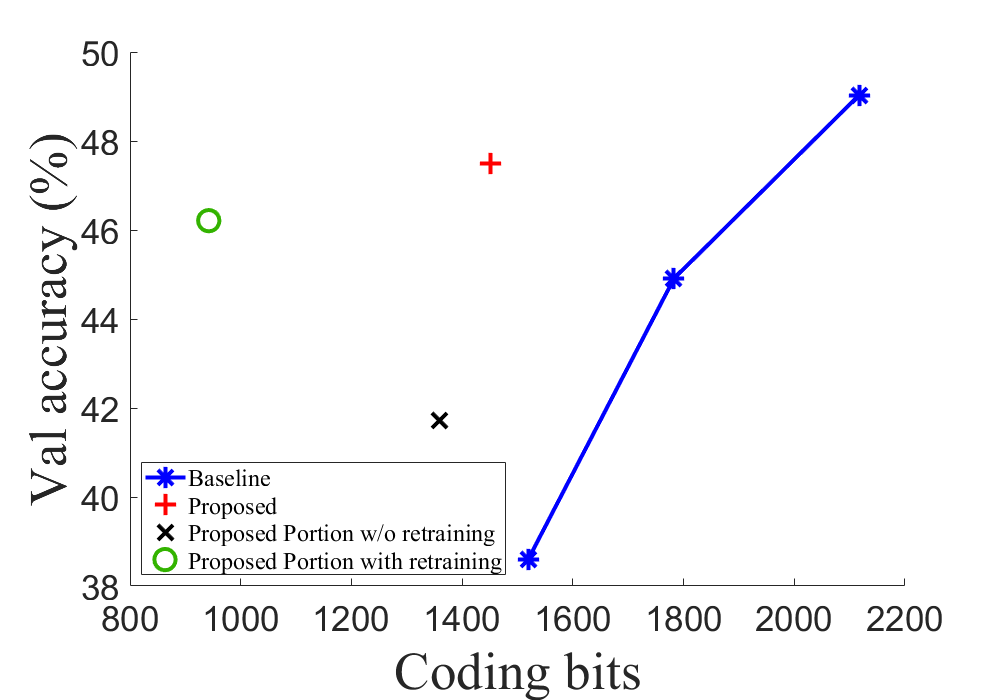}}
  \centerline{(a)}\medskip
\end{minipage}
\hfill
\begin{minipage}[b]{0.47\linewidth}
  \centering
  \centerline{\includegraphics[width=4.3cm]{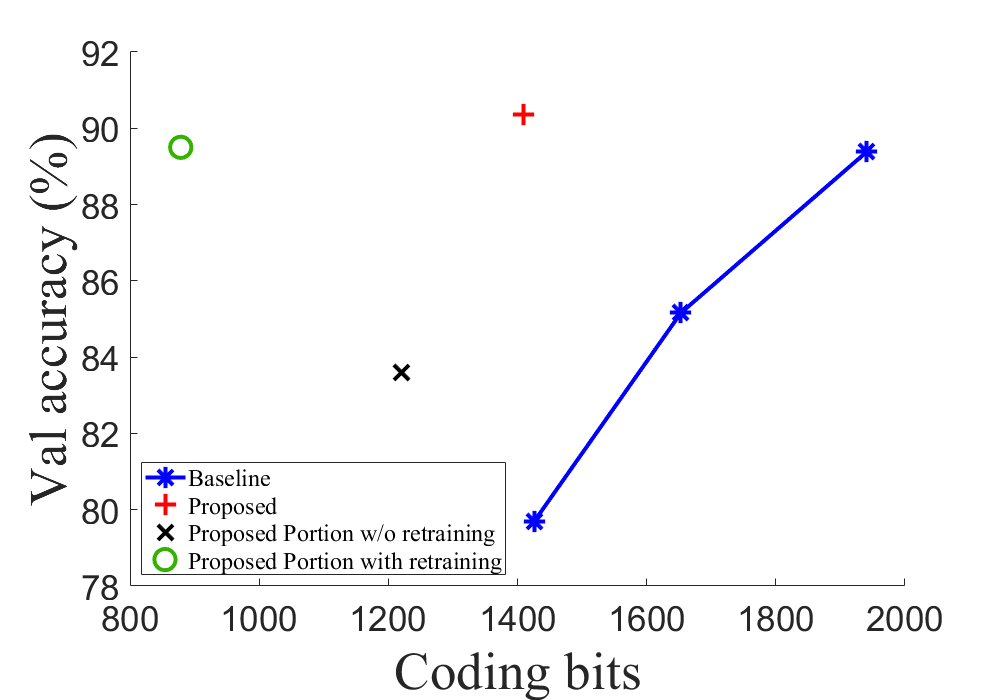}}
  \centerline{(b)}\medskip
\end{minipage}
\caption{The RA performance comparisons with baseline method, proposed method, and proposed method with a portion of descriptors (Proposed Portion with retraining and Proposed Portion w/o retraining) for two machine tasks. 
(a) The RA performance of the expression recognition task; (b) The RA performance of the face verification task. }
\label{Fig3}
\end{figure}

When considering facial expression recognition or face verification only, we can even remove a certain portion of the descriptors as not all components are necessary, due to the favourable interpretability of the 3D descriptors. Therefore, the RA performance of the expression recognition and face verification tasks can be evaluated with only the required descriptors transmitted, while other evaluation settings are maintained. Moreover, the descriptor compression network can be retrained with the specified portion, or retained to be the one with all descriptor components used for training. For the evaluation setting without retraining, the $\boldsymbol{\omega}$ contains the required descriptors of specific tasks, and other descriptors are padded with zeros. For the evaluation setting with retraining, the new compression networks are retrained with only the required descriptors taken as input. The other training settings are retained. As shown in Fig.~\ref{Fig3}, the bit consumption with a portion of the descriptors (w/o retraining) can be reduced compared to the case that all the descriptors are coded. On the other hand, when considering the retraining, significant bits saving with better accuracy in both tasks can be achieved. These experiments provide useful evidence that the proposed method could be used with partial descriptors and demonstrate the potential that the proposed framework could provide the reasonable capability to adapt to different applications with promising performance.

\section{Conclusions and Envisions}

We make one of the first attempts to establish the semantic face compression paradigm, targeting the problem of virtual avatar face communication in the metaverse. The unique challenge is that the facial information is not physically perceived in metaverse, as the receiver (e.g., virtual avatar) is built based on the AI system. We lay out a feasible solution based on compact 3D descriptors, which has shown many advantages towards this unique task. Experimental results validate the potential of the proposed scheme, thereby establishing face semantic communication in the metaverse as a novel paradigm in its own right. 

It is envisioned that the current work can inspire future research works that explore the communication of virtual avatars in metaverse beyond face information. For example, extension of the proposed scheme to virtual avatar body is straightforward, while more studies regarding how the semantic is utilized in a different scenario shall be carefully explored. Moreover, how the semantic information for face can be better represented to undertake more intelligent tasks shall also be explored. Another promising direction is to systematically investigate the performance evaluation methodology of semantic communication of virtual avatars in the metaverse. This poses new challenges to both the quality evaluation and semantic communication research, and opens up new space for future exploration.

\appendices

\ifCLASSOPTIONcaptionsoff
  \newpage
\fi

\bibliographystyle{IEEEtran}
\bibliography{refs}
 
\end{document}